\documentclass{article}

\usepackage[preprint,nonatbib]{neurips_2023}

\usepackage[utf8]{inputenc} 
\usepackage[T1]{fontenc}    
\usepackage{hyperref}       
\usepackage{url}            
\usepackage{booktabs}       
\usepackage{amsfonts}       
\usepackage{nicefrac}       
\usepackage{microtype}      
\usepackage{xcolor}         

\usepackage{appendix}
\usepackage{listings}
\usepackage{booktabs}
\usepackage{adjustbox}
\usepackage[accsupp]{axessibility}  
\usepackage{algorithm}
\usepackage{algorithmic}
\usepackage{wrapfig}
\usepackage{xcolor}
\usepackage{enumitem}
\usepackage{mathtools}
\usepackage{subfigure}
\usepackage{multirow}
\usepackage{multicol}
\usepackage{adjustbox}
\usepackage{enumitem}
\setitemize{itemsep=0pt,topsep=0pt,parsep=0pt,partopsep=0pt}
\input{Definitions.tex}
\title{Enhancing Detail Preservation for Customized Text-to-Image Generation:\\ A Regularization-Free Approach}

\author{%
Yufan Zhou $^1$,
~
Ruiyi Zhang $^2$,
~
Tong Sun $^2$,
~
Jinhui Xu $^1$
\\
$^1$ State University of New York at Buffalo~~~~ 
$^2$ Adobe Research~~~~
\\
{\tt\small \{yufanzho, jinhui\}@buffalo.edu}, ~~~~
{\tt\small \{ruizhang, tsun\}@adobe.com}~~~~ 
}

\begin{document}

\maketitle

\begin{abstract}
    Recent text-to-image generation models have demonstrated impressive capability of generating text-aligned images with high fidelity. However, generating images of novel concept provided by the user input image is still a challenging task. To address this problem, researchers have been exploring various methods for customizing pre-trained text-to-image generation models. Currently, most existing methods for customizing pre-trained text-to-image generation models involve the use of regularization techniques to prevent over-fitting. While regularization will ease the challenge of customization and leads to successful content creation with respect to text guidance, it may restrict the model capability, resulting in the loss of detailed information and inferior performance. In this work, we propose a novel framework for customized text-to-image generation without the use of regularization. Specifically, our proposed framework consists of an encoder network and a novel sampling method which can tackle the over-fitting problem without the use of regularization. With the proposed framework, we are able to customize a large-scale text-to-image generation model within half a minute on single GPU, with only one image provided by the user. We demonstrate in experiments that our proposed framework outperforms existing methods, and preserves more fine-grained details. 
\end{abstract}

\section{Introduction}
Text-to-image generation is a research topic that has been explored for years~\cite{tao2021dfgan,xu2018attngan,zhang2021cross,Zhou2022Lafite2FT,zhou2021lafite,zhu2019dm}, with remarkable progresses recently. Nowadays, researchers are able to perform zero-shot text-to-image generation with arbitrary text input by training large-scale models on web-scale datasets. 
Starting from DALL-E~\cite{ramesh2021zero} and CogView~\cite{ding2021cogview}, numerous methods have been proposed~\cite{chang2023muse,ding2022cogview2,gafni2022make,ramesh2022hierarchical,rombach2021high,saharia2022photorealistic,yu2022scaling,zhou2022shifted}, leading to impressive capability in generating text-aligned images of high resolution with exceptional fidelity. Besides text-to-image generation, these large-scale models also have huge impacts on many other applications including image manipulation~\cite{avrahami2022blended,hertz2022prompt} and video generation~\cite{ho2022imagenvideo,singer2022makeavideo}.

Although aforementioned large-scale text-to-image generation models are able to perform text-aligned and creative generation, they may face difficulties in generating novel and unique concepts~\cite{gal2022textualinversion} specified by users. Thus, researchers have exploited different methods in customizing pre-trained text-to-image generation models. For instance, \cite{kumari2022multi,ruiz2022dreambooth} propose to fine-tune the pre-trained generative models with few samples, where different regularization methods are applied to prevent over-fitting. \cite{gal2022textualinversion,gal2023designing,wei2023elite} propose to encode the novel concept of user input image in a word embedding, which is obtained by an optimization method or from an encoder network. All these methods lead to customized generation for the novel concept, while satisfying additional requirements described in arbitrary user input text. 

Despite these progresses, recent research also makes us 
suspect that the use of regularization may potentially restrict the capability of customized generation, leading to the information loss of fine-grained details. In this paper, we propose a novel framework called \textit{\text{ProFusion}}, which consists of an encoder called \textit{\textbf{Pro}mptNet} and a novel sampling method called \textit{\textbf{Fusion} Sampling}.
Different from previous methods, our ProFusion does not require any regularization, the potential over-fitting problem can be tackled by the proposed Fusion Sampling method at inference, which saves training time as there is no need to tune the hyper-parameters for regularization method. 
Our main contributions can be summarized as follows:

\begin{itemize}[nosep]
    \item We propose ProFusion, a novel framework for customized generation. Given single testing image containing a unique concept, the proposed framework can generate customized output for the unique concept and meets additional requirement specified in arbitrary text. Only about 30 seconds of fine-tuning on single GPU is required;
    \item The proposed framework does not require any regularization method to prevent over-fitting, which significantly reduces training time as there is no need to tune regularization hyper-parameters. The absence of regularization also allows the proposed framework to achieve enhanced preservation of fine-grained details;
    \item Extensive results,including qualitative, quantitative and human evaluation results, have demonstrated the effectiveness of the proposed ProFusion. Ablation studies are also conducted to better understand the components in the proposed framework;
\end{itemize}

\begin{figure}[t!]
    \centering
    \includegraphics[width=1.0\linewidth]{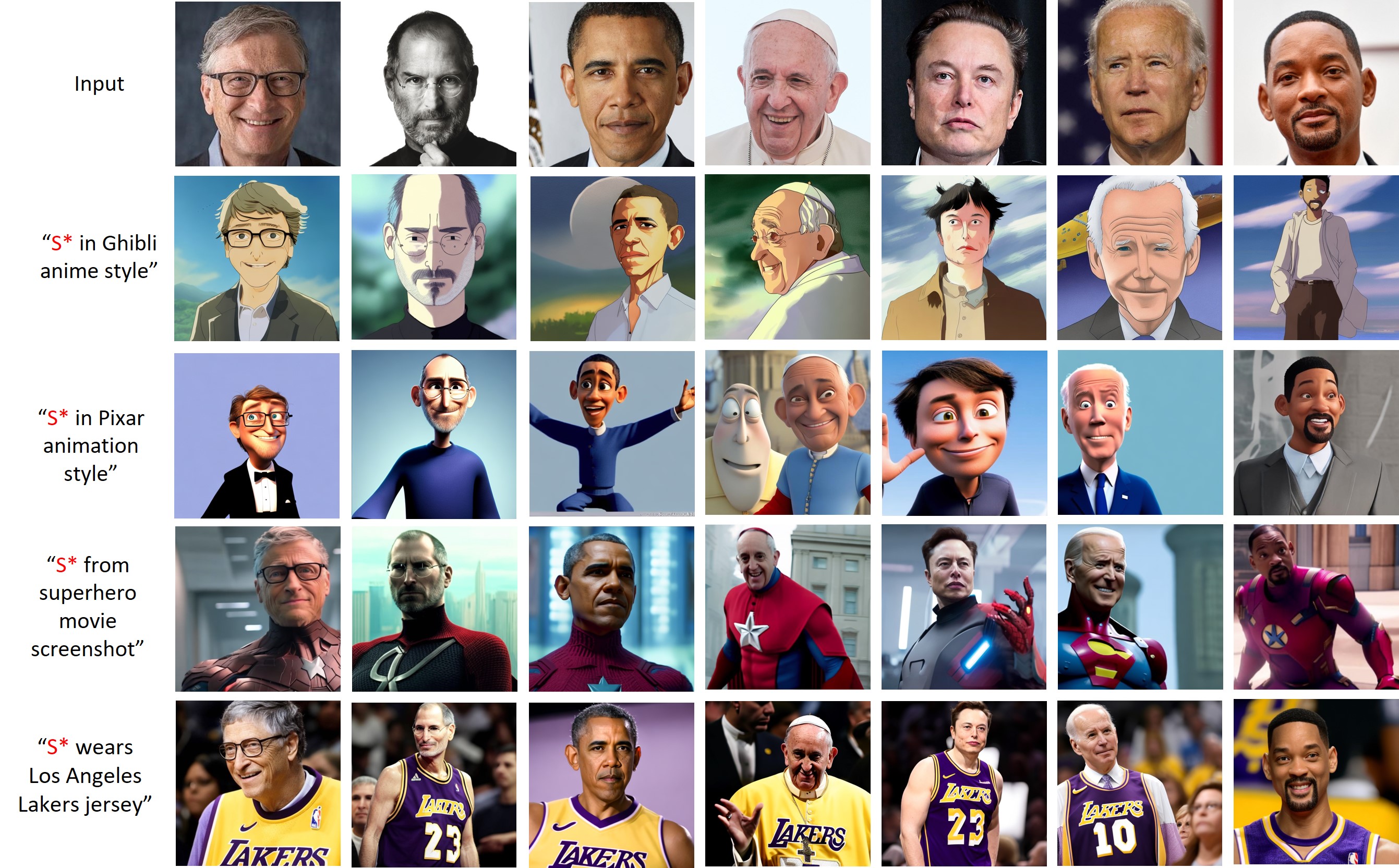}
    \caption{Customized generation with the proposed framework. Given only single testing image, we are able to perform customized generation which satisfies arbitrary specified requirements and preserves fine-grained details.}
    \vspace{-0.1in}
    \label{fig:main_results_1}
    \vspace{-0.1in}    
\end{figure}

\section{Methodology}
We now present our proposed ProFusion framework, which consists of a neural network called PromptNet and a novel sampling method called Fusion Sampling. Specifically, PromptNet is an encoder network which can generate word embedding $S^*$ conditioned on input image $\xb$, inside the input embedding space of the text encoder from Stable Diffusion 2. The major benefit of mapping $\xb$ into $S^*$ is that $S^*$ can be readily combined with arbitrary text to construct prompt for creative generation, \textit{e.g.}, "$S^*$ from a superhero movie screenshot"; Meanwhile, the Fusion Sampling is a sampling method leads to promising generation which meets the specified text requirements while maintaining fine-grained details of the input image $\xb$.

\begin{figure}[t!]
    \centering
    \includegraphics[width=1.0\linewidth]{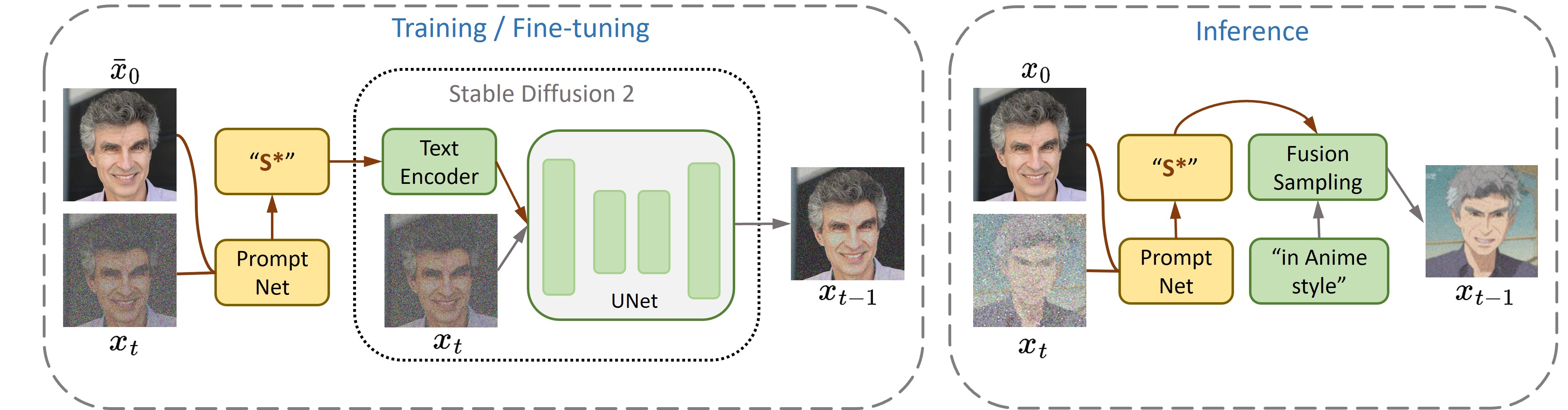}    
    \vspace{-0.1in}
    \caption{Illustration of the proposed framework.}
    \label{fig:framework}
\end{figure}

Our core idea is presented in Figure \ref{fig:framework}. The proposed PromptNet infers $S^*$ from an input image $\xb_0$ and current noisy generation $\xb_t$. Instead of using $\xb_0$, we can use $\bar{\xb}_0$ during the training of PromptNet, which denotes a different view of $\xb_0$ and can be obtained by data augmentation, \textit{e.g.}, resizing, rotation. The PromptNet is trained with diffusion loss:
\begin{align}\label{eq:loss_textual_inversion}
    L_{\text{Diffusion}} = \mathbb{E}_{\xb, \yb(S^*), t, \epsilonb \sim \mathcal{N}(\mathbf{0}, \mathbf{I})}\left[ \Vert \epsilonb - \epsilonb_{\thetab}(\xb_t, \yb(S^*), t) \Vert^2_2\right],
\end{align}
where $\yb(S^*)$ denotes the constructed prompt containing $S^*$, e.g. "A photo of $S^*$".

Existing works~\cite{gal2022textualinversion,gal2023designing} use similar idea to obtain $S^*$. However, regularization are often applied in these works. For instance, 
E4T~\cite{gal2023designing} proposes to use an encoder to generate $S^*$, which is optimized with
\begin{align}\label{eq:regularized_loss_texutal_inversion}
    L = L_{\text{Diffusion}} + \lambda \Vert S^* \Vert_2^2,
\end{align}
where the $L_2$ norm of $S^*$ is regularized. Similarly, Textual Inversion~\cite{gal2022textualinversion} proposes to directly obtain $S^*$ by solving 
\[
S^* = \text{argmin}_{S^\prime} L_{\text{Diffusion}} + \lambda \Vert S^\prime - S\Vert^2_2
\]
with optimization method, where $S$ denotes a coarse embedding\footnote{let $S^*$ be a target embedding for a specific human face image, $S$ can be set to be the embedding of text "face".}.

\begin{figure}[t!]
    \centering
    \includegraphics[width=0.7\linewidth]{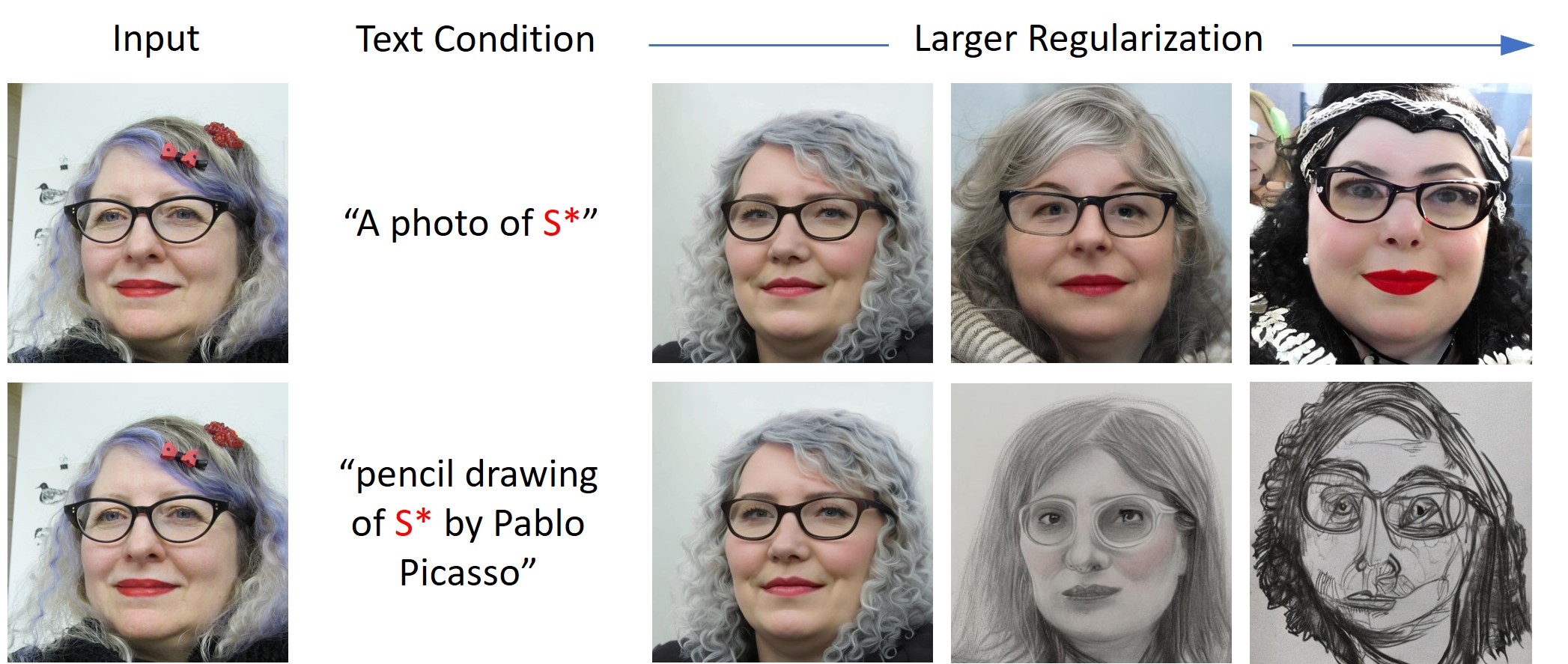}    
    \caption{The performance of customized generation is impacted by the level of regularization.}
    \vspace{-0.1in}
    \label{fig:regularized}
\end{figure}

In this work, we argue that although the use of regularization will ease the challenge and enables successful content creation with respect to testing text. It also leads to the loss of detailed information, resulting in inferior performance. To verify this argument, we conduct a simple experiment on FFHQ dataset~\cite{karras2019style}. We train several encoders with different levels of regularization by selecting different $\lambda$ in \eqref{eq:regularized_loss_texutal_inversion}. After training, we test their capability by classifier-free sampling~\cite{ho2021classifier} with different prompts containing resulting $S^*$. The results are shown in Figure \ref{fig:regularized}, from which we can find that smaller regularization leads to less information loss, which results in better preservation of details. However, the information could be too strong to prevent creative generation with respect to user input text. Meanwhile, large regularization leads to successful content creation, while fails to capture details of the input image, resulting in unsatisfactory results.

A consequent question is, \textbf{is it possible to perform successful customized generation using $S^*$ obtained without regularization so that the details from original image can be well-preserved?} To answer this question, 
we propose a novel sampling method called Fusion Sampling.

\subsection{Fusion Sampling}
Given a PromptNet pre-trained without regularization which can map input image $\xb_0$ into word embedding $S^*$, our goal is to successfully perform customized generation which  preserves details of $\xb_0$, and meets the requirements specified in arbitrary prompt containing $S^*$.

The task can be formulated as a conditional generation task with conditions $S^*$ and $C$, where $C$ denotes arbitrary user input text. 
We start from the most commonly used classifier-free sampling~\cite{ho2021classifier}.
To sample $\xb_{t-1}$ given current noisy sample $\xb_t$ and conditions $\left[S^*, C\right]$, the diffusion model first outputs the predictions of conditional noise $\epsilonb_{\thetab}(\xb_t, S^*, C)$ and unconditional noise $\epsilonb_{\thetab}(\xb_t)$. Then an updated prediction (with hyper-parameter $\omega$)
\begin{align}\label{eq:ddpm_eps}
    \tilde{\epsilonb}_{\thetab}(\xb_t, S^*, C) = (1+\omega)\epsilonb_{\thetab}(\xb_t, S^*, C) - \omega \epsilonb_{\thetab}(\xb_t), 
\end{align}
will be used in different sampling strategies~\cite{ho2020denoising,karraselucidating,song2020denoising,songscore}. 

In customized generation, the reason that vanilla classifier-free  sampling does not work without regularization is that, information from $S^*$ can become too strong without regularization. As a result, $\epsilonb_{\thetab}(\xb_t, S^*, C)$ will degenerate to $\epsilonb_{\thetab}(\xb_t, S^*)$ and information of $C$ will be lost. Thus, we need to propose a new sampling method, to produce a new prediction for $\tilde{\epsilonb_{\thetab}}(\xb_t, S^*, C)$  which is enforced to be conditioned on both $S^*$ and $C$.

\paragraph{Sampling with independent conditions}
We begin by assuming that $S^*$ and $C$ are independent. According to \cite{ho2021classifier}, we know that
\begin{align}\label{eq:eps_log}
    \epsilonb_{\thetab}(\xb_t, S^*, C) = -\sqrt{1-\bar{\alpha}_t}\nabla \log p(\xb_t\vert S^*, C),
\end{align}
where $\bar{\alpha}_t$ is a hyper-parameter as defined in \cite{ho2020denoising}. By \eqref{eq:eps_log} and Bayes' Rule,  we can re-write \eqref{eq:ddpm_eps} as
\begin{align}
     \tilde{\epsilonb}_{\thetab}(\xb_t, S^*, C) = \epsilonb_{\thetab}(\xb_t) - (1 + \omega) \sqrt{1-\bar{\alpha}_t}\nabla \log p (S^*, C\vert \xb_t). 
\end{align}
Since we assume that $S^*, C$ are independent, we can further re-write the above as
\begin{align*}
     \tilde{\epsilonb}_{\thetab}(\xb_t, S^*, C) & = \epsilonb_{\thetab}(\xb_t) - (1 + \omega) \sqrt{1-\bar{\alpha}_t}\nabla \log p (S^*\vert \xb_t) - (1 + \omega) \sqrt{1-\bar{\alpha}_t}\nabla \log p (C\vert \xb_t) \nonumber \\
     & = \epsilonb_{\thetab}(\xb_t)  + (1 + \omega) \{\epsilonb_{\thetab}(\xb_t, S^*) - \epsilonb_{\thetab}(\xb_t)\} + (1 + \omega) \{\epsilonb_{\thetab}(\xb_t, C) - \epsilonb_{\thetab}(\xb_t)\}. 
\end{align*}
We re-write it as 
\begin{align}\label{eq:independent_fusion}
     \tilde{\epsilonb}_{\thetab}(\xb_t, S^*, C) = \epsilonb_{\thetab}(\xb_t)  + (1 + \omega_1) \{\epsilonb_{\thetab}(\xb_t, S^*) - \epsilonb_{\thetab}(\xb_t)\} + (1 + \omega_2) \{\epsilonb_{\thetab}(\xb_t, C) - \epsilonb_{\thetab}(\xb_t)\} 
\end{align}
for more flexibility. \eqref{eq:independent_fusion} can be readily extended to more complicated scenarios, where a list of conditions $\{S_1^*, S_2^*,...,S_k^*, C\}$ are provided. The corresponding $\tilde{\epsilonb_{\thetab}}(\xb_t, \{S_i^*\}_{i=1}^k, C)$ is 
\begin{align*}
     \tilde{\epsilonb}_{\thetab}(\xb_t, \{S_i^*\}_{i=1}^k, C) = \epsilonb_{\thetab}(\xb_t)  + \sum_{i=1}^k (1 + \omega_i) \{\epsilonb_{\thetab}(\xb_t, S_i^*) - \epsilonb_{\thetab}(\xb_t)\} + (1 + \omega_C) \{\epsilonb_{\thetab}(\xb_t, C) - \epsilonb_{\thetab}(\xb_t)\}. 
\end{align*}
\paragraph{Fusion Sampling with dependent conditions}
One major drawback of \eqref{eq:independent_fusion} is that the independence does not always hold in practice. As we will show in later experiment, assuming $S^*$ and $C$ to be independent can lead to inferior generation.

To solve this problem, we propose Fusion Sampling, which consists of two stages at each timestep $t$: a \textbf{fusion stage} which encodes information from both $S^*$ and $C$ into $\xb_t$ with an updated $\tilde{\xb}_t$, and a \textbf{refinement stage} which predicts $\xb_{t-1}$ based on Equation \eqref{eq:independent_fusion}. 
The proposed algorithm is presented in Algorithm \ref{algo:fusion_sampling}. Sampling with independent conditions can be regarded as a special case of Fusion Sampling with $m=0$. 
In practice, $m=1$ works well, thus we set $m=1$ in all our experiments.
\begin{algorithm}[t!]
    \caption{Fusion Sampling at Timestep t}\label{algo:fusion_sampling}
    \begin{algorithmic}[1]
        \STATE {\bfseries Require: Conditions $S^*$ and $C$, a noisy sample $\xb_t$, a pre-trained diffusion model $\epsilonb_{\thetab}$, hyper-parameters  $0<\sigma_t, 0\leq\gamma \leq 1$.}     
        \STATE{Set $\tilde{\xb}_t = \xb_t$}
        \STATE{\color{blue} // Fusion Stage}        
        \FOR{i = 1, ..., m}
            \STATE{Generate $\tilde{\epsilonb_{\thetab}}(\tilde{\xb}_t, \gamma S^*, C)$ by \eqref{eq:ddpm_eps}.}
            \STATE{Generate predicted sample $\tilde{\xb}_0 = \dfrac{\tilde{\xb}_t - \sqrt{1 - \bar{\alpha_t}}\tilde{\epsilonb_{\thetab}}(\tilde{\xb}_t, \gamma S^*, C)}{\sqrt{\bar{\alpha}_t}}$. }
            \STATE{Inject fused information into $\tilde{\xb}_{t-1}$ by sampling $\tilde{\xb}_{t-1}\sim q(\tilde{\xb}_{t-1} \vert \tilde{\xb}_t, \tilde{\xb}_0)$.}
            \IF{Use refinement stage}
                \STATE Inject fused information into $\tilde{\xb}_{t}$ by sampling $\tilde{\xb}_t \sim q(\tilde{\xb}_t \vert \tilde{\xb}_{t-1}, \tilde{\xb}_0)$.
            \ELSE
                \STATE Return $\xb_{t-1} = \tilde{\xb}_{t-1}$.
            \ENDIF
        \ENDFOR
        \STATE{\color{blue} // Refinement Stage}
        \IF{Use refinement stage}
            \STATE Generate $\tilde{\epsilonb_{\thetab}}(\tilde{\xb}_t, S^*, C) $ by \eqref{eq:independent_fusion} and perform classifier-free sampling step. Return $\xb_{t-1}$.
        \ENDIF
    \end{algorithmic}
\end{algorithm}

The remaining challenge in Algorithm \ref{algo:fusion_sampling} is to sample $\tilde{\xb}_{t-1}\sim q(\tilde{\xb}_{t-1} \vert \tilde{\xb}_t, \tilde{\xb}_0)$ and $\tilde{\xb}_t \sim q(\tilde{\xb}_t \vert \tilde{\xb}_{t-1}, \tilde{\xb}_0)$. We take Denoising Diffusion Implicit Models (DDIM)~\cite{song2020denoising} as an example, while the following derivation can be extended to other diffusion models.
Let $\mathbf{I}$ be the identity matrix, $\sigma_t$ denotes a hyper-parameter controlling randomness. In DDIM, we have
\begin{align}\label{eq:ddim_q_x_t}
    q(\tilde{\xb}_t \vert \tilde{\xb}_0) = \mathcal{N}(\tilde{\xb}_t; \sqrt{\bar{\alpha}_t} \tilde{\xb}_0, (1 - \bar{\alpha}_t) \mathbf{I})
\end{align}
and
\begin{align}
    q(\tilde{\xb}_{t-1} \vert \tilde{\xb}_t, \tilde{\xb}_0) = \mathcal{N}(\tilde{\xb}_{t-1}; \sqrt{\bar{\alpha}_{t-1}}\tilde{\xb}_0 + \sqrt{1 - \bar{\alpha}_{t-1} - \sigma_t^2} \dfrac{\tilde{\xb}_t - \sqrt{\bar{\alpha}_t}\tilde{\xb}_0}{\sqrt{1 - \bar{\alpha}_t}}, \sigma_t^2\mathbf{I}).
\end{align}
By the property of Gaussian distributions~\cite{bishop2006PRML}, we know that
\begin{align}
     q(\tilde{\xb}_t \vert \tilde{\xb}_{t-1}, \tilde{\xb}_0) = \mathcal{N}(\tilde{\xb}_t; \Sigmab(A^T L(\tilde{\xb}_{t-1} - b) + B\mub), \Sigmab)
\end{align}
where
\begin{align*}
    \Sigmab &= \dfrac{(1 - \bar{\alpha}_t)\sigma_t^2}{1 - \bar{\alpha}_{t-1}} \mathbf{I}, 
    ~~~~~ \mub = \sqrt{\bar{\alpha}_t} \tilde{\xb}_0, 
    ~~~~~ b = \sqrt{\bar{\alpha}_{t-1}}\tilde{\xb}_0 - \dfrac{\sqrt{\bar{\alpha}_t(1 - \bar{\alpha}_{t-1} - \sigma_t^2)}}{\sqrt{1 - \bar{\alpha}_t}} \tilde{\xb}_0 \\
    A & = \dfrac{\sqrt{1 - \bar{\alpha}_{t-1} - \sigma_t^2}}{\sqrt{1 - \bar{\alpha}_t}} \mathbf{I}, 
    ~~~~~L = \dfrac{1}{\sigma_t^2} \mathbf{I}, 
    ~~~~~ B = \dfrac{1}{1 - \bar{\alpha}_t}\mathbf{I}
\end{align*}
which leads to 
\begin{align}
    \tilde{\xb}_t =& \dfrac{\sqrt{(1 - \bar{\alpha}_t)(1 - \bar{\alpha}_{t-1} - \sigma_t^2)}}{1 - \bar{\alpha}_{t-1}} \tilde{\xb}_{t-1} + \dfrac{(1 - \bar{\alpha}_t)\sigma_t^2}{1 - \bar{\alpha}_{t-1}} \zb \nonumber \\
    & + \dfrac{\tilde{\xb}_0}{1 - \bar{\alpha}_{t-1}}\{\sqrt{\bar{\alpha}_t}(1 - \bar{\alpha}_{t-1}) - \sqrt{\bar{\alpha}_{t-1}(1 - \bar{\alpha}_{t})(1 - \bar{\alpha}_{t-1} - \sigma_t^2)})\}, ~~~~ \zb \sim \mathcal{N}(\mathbf{0}, \mathbf{I}).
\end{align}

With further derivation, we can summarize a single update in fusion stage as:
\begin{align}\label{eq:update_xt}
    \tilde{\xb}_{t} \leftarrow \tilde{\xb}_t - \dfrac{\sigma_t^2 \sqrt{1 - \bar{\alpha}_t}}{1 - \bar{\alpha}_{t-1}} \tilde{\epsilonb}_{\thetab}(\tilde{\xb}_t, \gamma S^*, C) + \dfrac{\sqrt{(1 - \bar{\alpha}_t)(2 - 2\bar{\alpha}_{t-1} - \sigma_t^2)}}{1 - \bar{\alpha}_{t-1}}\sigma_t \zb, ~~\zb\sim \mathcal{N}(\mathbf{0}, \mathbf{I}).
\end{align}
\begin{remark}
    Recall $\tilde{\epsilonb}_{\thetab}(\tilde{\xb}_t, \gamma S^*, C) = -\sqrt{1 - \bar{\alpha}_t}\nabla \log \tilde{p}_{\omega}(\tilde{\xb}_t\vert \gamma S^*, C)$~\cite{ho2021classifier}, we can re-write \eqref{eq:update_xt} as 
    \begin{align}\label{eq:actually_update_xt}
        \tilde{\xb}_{t} \leftarrow \tilde{\xb}_t + \dfrac{\sigma_t^2 (1 - \bar{\alpha}_t)}{1 - \bar{\alpha}_{t-1}} \nabla \log \tilde{p}_{\omega}(\tilde{\xb}_t\vert \gamma S^*, C) + \dfrac{\sqrt{(1 - \bar{\alpha}_t)(2 - 2\bar{\alpha}_{t-1} - \sigma_t^2)}}{1 - \bar{\alpha}_{t-1}}\sigma_t \zb.
    \end{align}
    From \eqref{eq:actually_update_xt}, we can conclude that our fusion stage is actually an gradient-based optimization method similar to Langevin dynamics~\cite{welling2011bayesian}.
    Compared to Langevin dynamics which is
    \begin{align}
            \tilde{\xb}_{t} \leftarrow \tilde{\xb}_t + \lambda \nabla \log \tilde{p}_{\omega}(\tilde{\xb}_t\vert \gamma S^*, C) + \sqrt{2\lambda} \zb.
    \end{align}
    with $\lambda$ being the step size, \eqref{eq:actually_update_xt} has less randomness, because
    \begin{align*}
         \dfrac{(1 - \bar{\alpha}_t)(2 - 2\bar{\alpha}_{t-1} - \sigma_t^2)\sigma_t^2}{(1 - \bar{\alpha}_{t-1})^2} \leq  \dfrac{2\sigma_t^2 (1 - \bar{\alpha}_t)}{1 - \bar{\alpha}_{t-1}}. 
    \end{align*}
\end{remark}
\begin{remark}
    If we set the DDIM hyper-parameter to be $\sigma_t = \sqrt{1 - \bar{\alpha}_{t-1}}$, then \eqref{eq:update_xt} becomes 
    \[
    \tilde{\xb}_t \leftarrow \tilde{\xb}_t - \sqrt{1 - \bar{\alpha}_t} \tilde{\epsilonb}(\tilde{\xb}_t, \gamma S^*, C) + \sqrt{1 - \bar{\alpha}_t}\zb, ~~~\zb\sim \mathcal{N}(\mathbf{0}, \mathbf{I})
    \]
    which is equivalent to sampling $\tilde{\xb}_t$ using \eqref{eq:ddim_q_x_t} without sampling intermediate $\tilde{\xb}_{t-1}$ in our Algorithm \ref{algo:fusion_sampling}. Thus directly sampling $\tilde{\xb}_t$ using \eqref{eq:ddim_q_x_t} is a special case of our Fusion Sampling algorithm.
\end{remark}

\section{Experiments}

We conduct extensive experiments to evaluate the proposed framework. Specifically, we first pre-train a PromptNet on FFHQ dataset~\cite{karras2019style} on 8 NVIDIA A100 GPUs for 80,000 iterations with a batch size of 64, without any data augmentation.
Given a testing image, the PromptNet and all attention layers of the pre-trained Stable Diffusion 2 are fine-tuned for 50 steps with a batch size of 8. Only half a minute and a single GPU is required in fine-tuning such a customized generative model, indicating the efficiency of the proposed method, especially considering the impressive results we could obtain. Some more implementation details are provided in the Appendix.
Our code and pre-trained models will be publicly available at \url{https://github.com/drboog/ProFusion}. 

\begin{figure}[t!]
    \centering
    \includegraphics[width=0.95\linewidth]{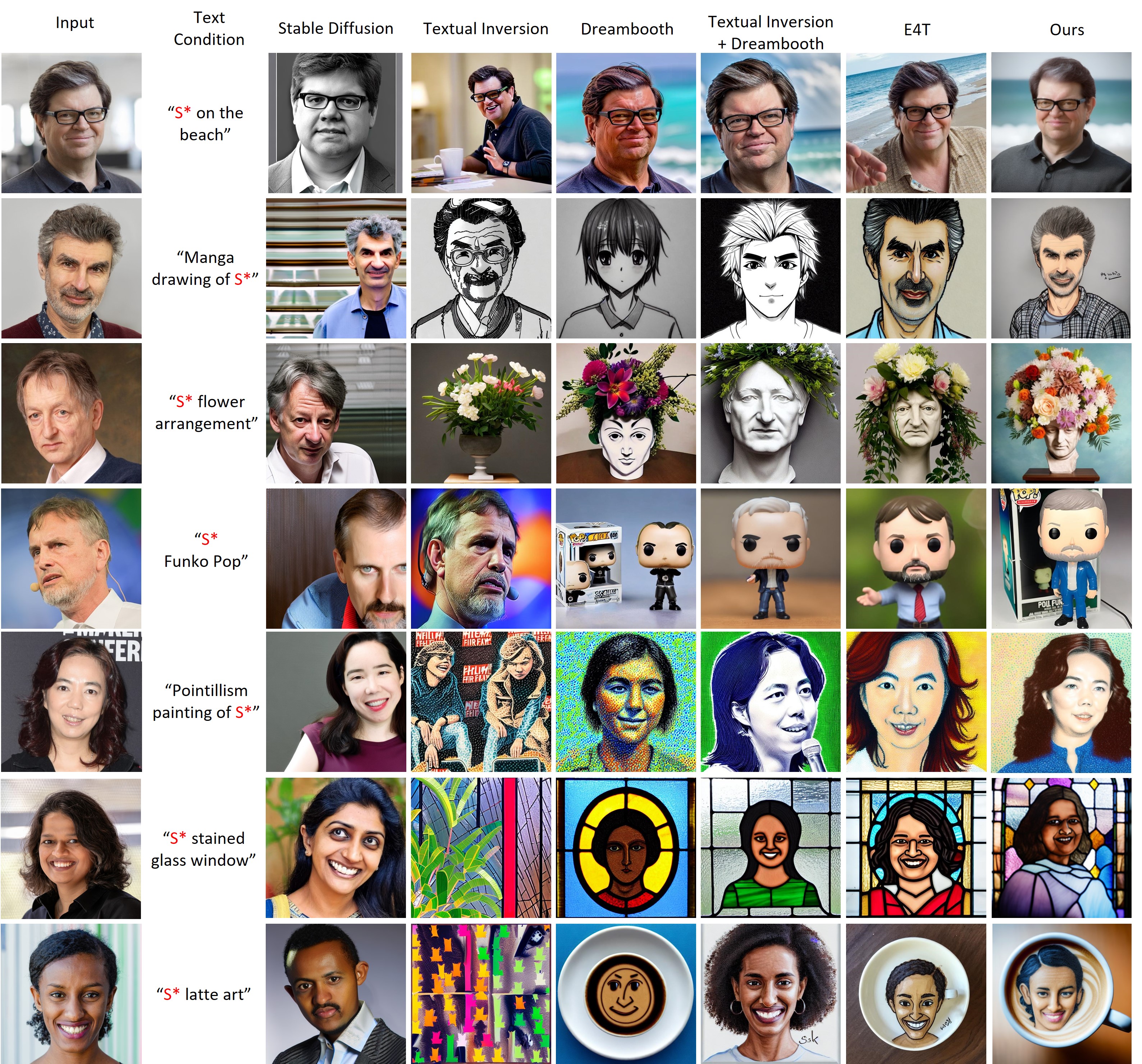}
    \vspace{-0.1in}    
    \caption{Comparison with baseline methods. Our proposed approach exhibits superior capability for preserving fine-grained details.}
    \label{fig:comparison}
    \vspace{-0.2in}
\end{figure}

\begin{figure}[t!]
    \centering
    \includegraphics[width=0.9\linewidth]{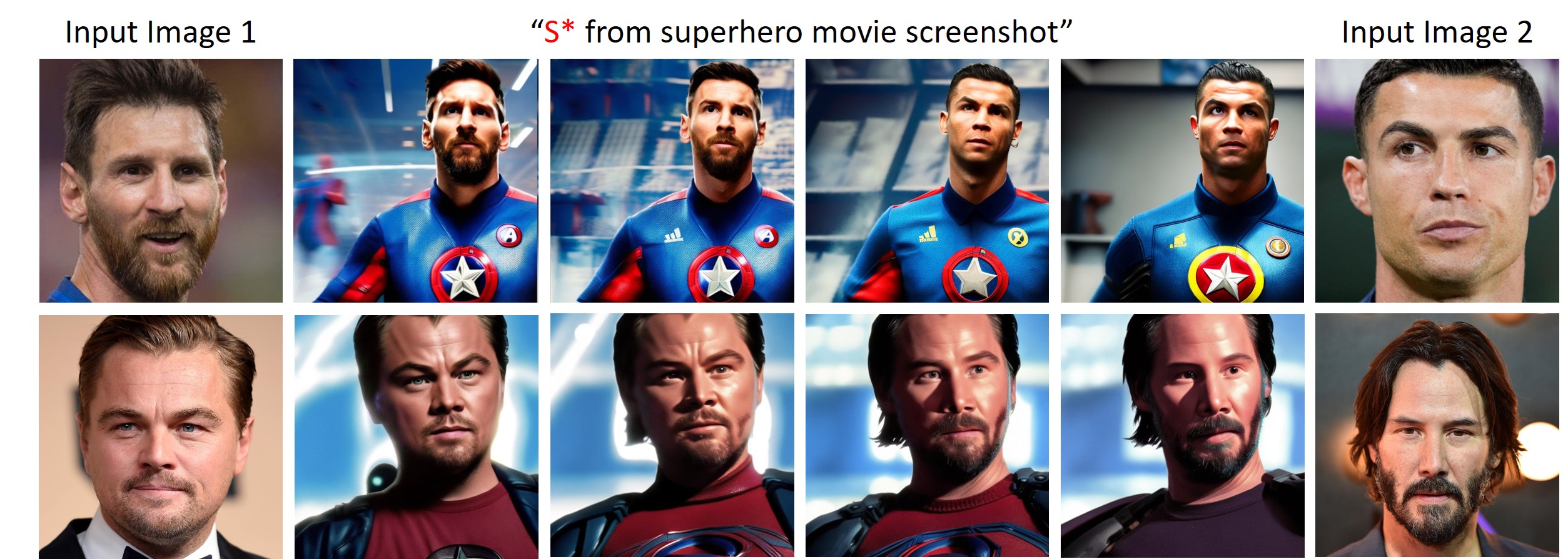}
    \vspace{-0.1in}
    \caption{The proposed framework enables generation conditioned on multiple input images and text. Creative interpolation can be performed.}
    \label{fig:interpolation}
    \vspace{-0.15in}    
\end{figure}

\subsection{Qualitative Results}
\begin{figure}[t!]
    \centering
    \includegraphics[width=0.7\linewidth]{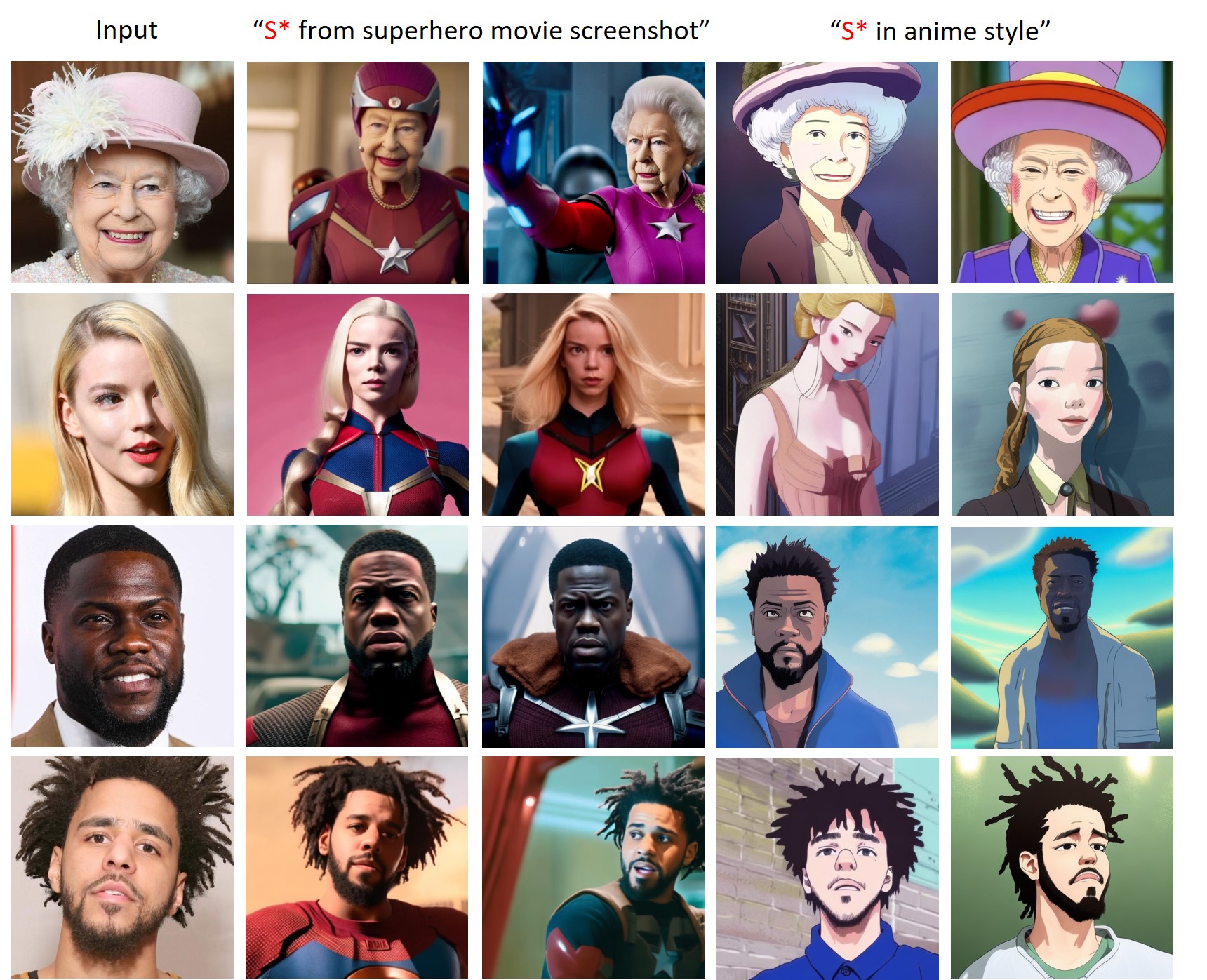}
    \vspace{-0.1in}
    \caption{Some results of customized generation with the proposed framework.}
    \label{fig:main_results_2}
    \vspace{-0.2in}
\end{figure}

Our main results are shown in Figure \ref{fig:main_results_1} and Figure \ref{fig:main_results_2}. From the results, we can see that the proposed framework effectively achieves customized generation which meets the specified text requirements while maintaining fine-grained details of the input image. More results are provided in the Appendix. As mentioned previously, our proposed framework is also able to perform generation conditioned on multiple images. We also provide these generated examples in Figure \ref{fig:interpolation}.

\begin{table}[t!]
    \centering
    \scalebox{0.7}{
    \begin{tabular}{lccccccccc}
        \toprule
        \multirow{2}{*}{Method} & \multicolumn{9}{c}{Pre-trained CLIP Models}\\
        & ViT-B/32 & ViT-B/16 & ViT-L/14 & ViT-L/14@336px & RN101 & RN50 & RN50$\times 4$& RN50$\times 16$& RN50$\times 64$\\
        \midrule
        Stable Diffusion 2 & 0.271 & 0.256 & 0.196 & 0.196 & 0.428 & 0.202 & 0.355 & 0.254 & 0.181 \\
        Textual Inversion & 0.257 & 0.251 & 0.197 & 0.201 & 0.426 & 0.195 & 0.350 & 0.247 & 0.173 \\
        DreamBooth & 0.283 & 0.267 & 0.205 & 0.210 & 0.434 & 0.209 & 0.363 & 0.260 & 0.187 \\
        E4T & 0.277 & 0.264 & 0.203 & 0.213 & 0.429 & 0.206 & 0.358 & 0.260 & 0.191\\
        \textbf{ProFusion (Ours)} & \textbf{0.293} & \textbf{0.283} & \textbf{0.225} & \textbf{0.229} & \textbf{0.446} & \textbf{0.223} & \textbf{0.374} & \textbf{0.279} & \textbf{0.202}\\
        \bottomrule
    \end{tabular}
    }
    \caption{Similarity ($\uparrow$) between generated example and input text.}    %
    \vspace{-0.2in}
    \label{tab:prompt_similarity}
\end{table}

Following \cite{gal2023designing}, we then compare proposed framework with several baseline methods including Stable Diffusion\footnote{The results of Stable Diffusion is obtained by directly feeding corresponding researcher's name and text requirements into the pre-trained text-to-image generation model. }~\cite{rombach2021high}, Textual Inversion~\cite{gal2022textualinversion}, DreamBooth~\cite{ruiz2022dreambooth}, E4T~\cite{gal2023designing}. The qualitative results are presented in Figure \ref{fig:comparison}, where the results of related methods are directly taken from ~\cite{gal2023designing}. From the comparison we can see that our framework results in better preservation of fine-grained details.

\subsection{Quantitative Results}
We also evaluate our methods and baseline methods quantitatively. Specifically, we utilize different pre-trained CLIP models~\cite{radford2021learning} to calculate the image-prompt similarity between the generated image and input text. The results are shown in Table \ref{tab:prompt_similarity}, our ProFusion obtains higher image-prompt similarity on all CLIP models, indicating better prompt-adherence and edit-ability..

We then calculate the identity similarity between the generated image and input image, which is cosine similarity computed using features extracted by pre-trained face recognition models. The identity similarity is also evaluated across different pre-trained models~\cite{deng2019arcface,kim2022adaface,vggface,schroff2015facenet,serengil2020lightface,serengil2021lightface,serengil2023db,taigman2014deepface}. The results are shown in Table \ref{tab:identity_similarity}. In general, our ProFusion obtains higher similarity, indicating better identity preservation.

\subsection{Human Evaluation}
We then conduct human evaluation on Amazon Mechanical Turk (MTurk). 
The workers are presented with two generated images from different methods along with original image and text requirements. They are then tasked with indicating their preferred choice. More details are provided in the Appendix.
The results are shown in Figure \ref{fig:human_evaluation}, where we can find that our method obtains a higher preference rate compared to all other methods, indicating the effectiveness of our proposed framework.

\begin{table}[t!]
    \centering
    \scalebox{0.8}{
    \begin{tabular}{lcccccccc}
        \toprule
        \multirow{2}{*}{Method} & \multicolumn{8}{c}{Pre-trained Face Recognition Models}\\
        & VGG-Face & Facenet & Facenet512 & OpenFace & DeepFace & ArcFace & SFace & AdaFace\\
        \midrule
        Stable Diffusion 2 & 0.530 & 0.334 & 0.323 & 0.497 & 0.641 & 0.144 & 0.191 & 0.093\\
        Textual Inversion & 0.516 & 0.410 & 0.372 & 0.566 & 0.651 & 0.248 & 0.231 & 0.210 \\
        DreamBooth & 0.518 & 0.483 & 0.415 & 0.516 & 0.643 & 0.379 & 0.304 & 0.307 \\
        E4T &  0.677 & 0.596 & \textbf{0.621} & 0.660 & 0.732 & 0.454 & 0.398 & 0.426 \\
        \textbf{ProFusion (Ours)} &  \textbf{0.720} & \textbf{0.616} & 0.597 & \textbf{0.681} & \textbf{0.774} & \textbf{0.459} & \textbf{0.443} & \textbf{0.432} \\
        \bottomrule
    \end{tabular}
    }
    \caption{Similarity ($\uparrow$) between generated example and input image.}
    \vspace{-0.2in}        
    \label{tab:identity_similarity}
\end{table}

\begin{figure}[t!]
    \centering
    \subfigure[vs. Stable Diffusion]{\includegraphics[width=0.24\linewidth]{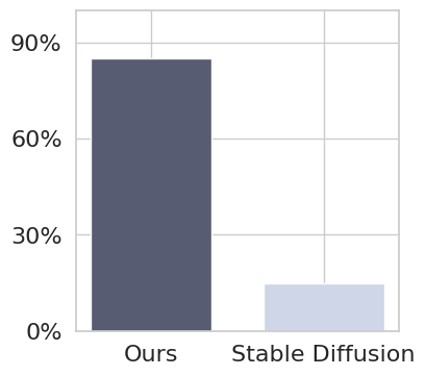}}  
    \subfigure[vs. Textual Inversion]{\includegraphics[width=0.243\linewidth]{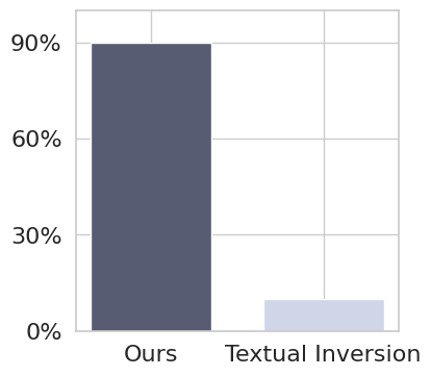}}  
    \subfigure[vs. DreamBooth]{\includegraphics[width=0.23\linewidth]{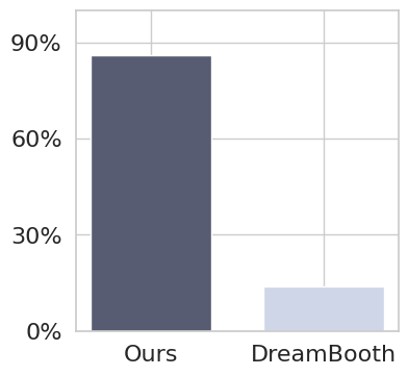}}  
    \subfigure[vs. E4T]{\includegraphics[width=0.23\linewidth]{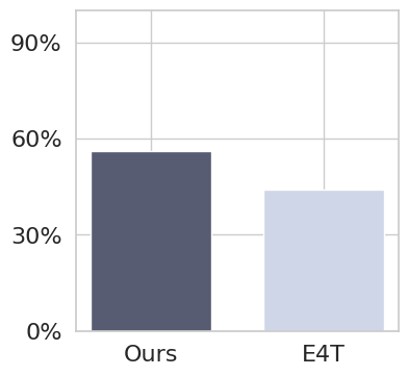}}  
    \vspace{-0.1in}
    \caption{Results of human evaluation.}
    \label{fig:human_evaluation}
    \vspace{-0.in}
\end{figure}

\begin{figure}[t!]
    \centering
    \includegraphics[width=0.7\linewidth]{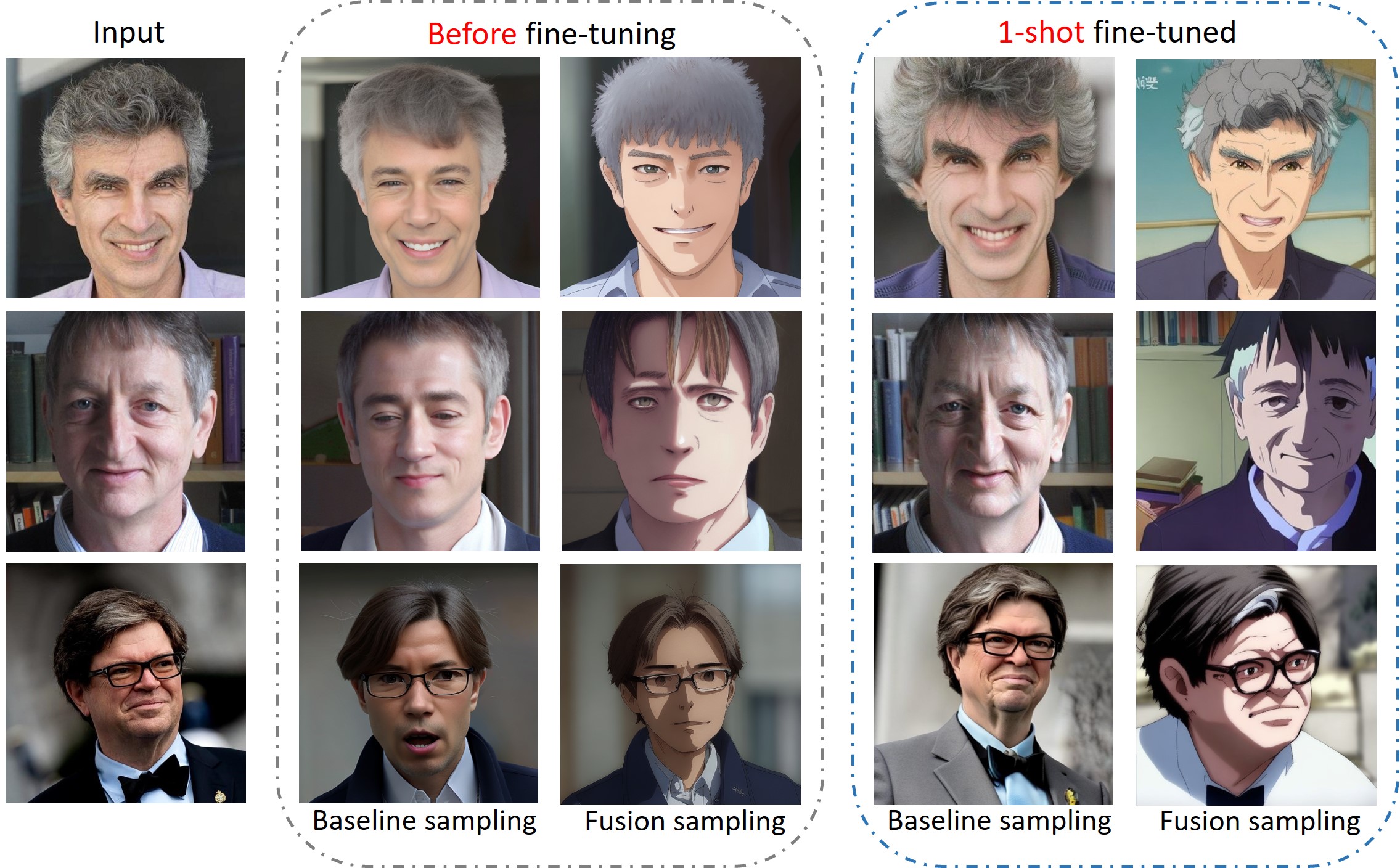}
    \vspace{-0.1in}    
    \caption{Examples with prompt "$\color{red}{S^*}$ in anime style", Fusion Sampling outperforms baseline.}
    \vspace{-0.2in}    
    \label{fig:before_after_finetuning}
\end{figure}

\subsection{Ablation Study}
We conduct several ablation studies to further investigate the proposed ProFusion. 

\begin{figure}[t!]
    \centering
    \includegraphics[width=0.7\linewidth]{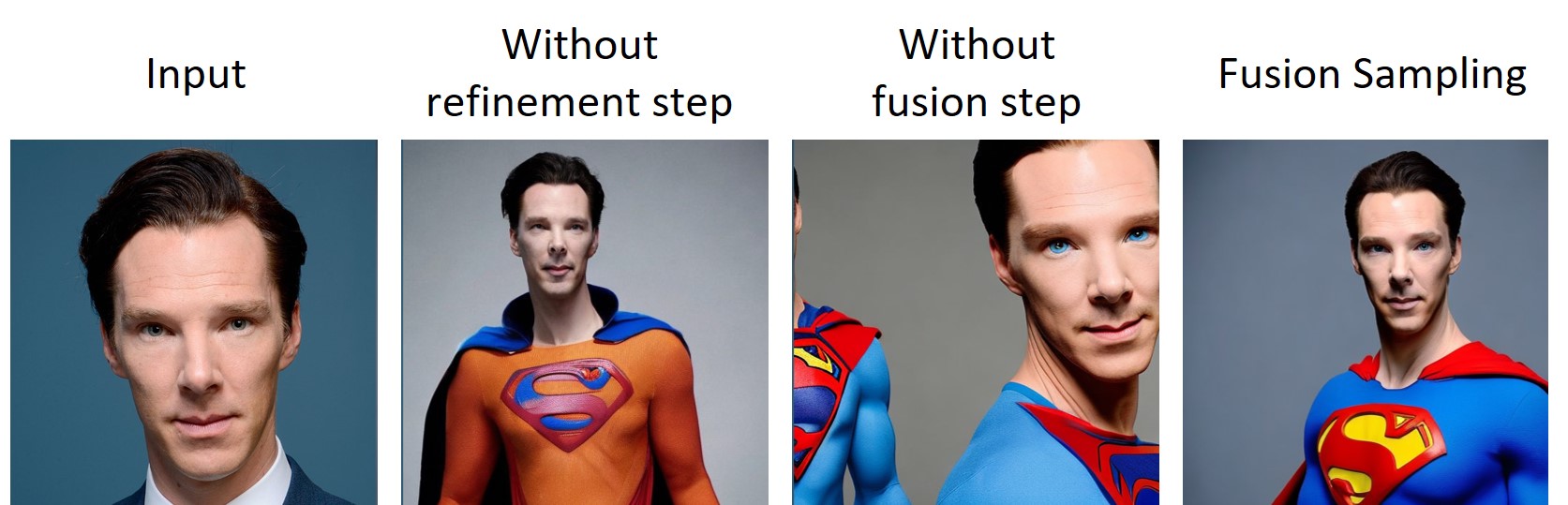}
    \vspace{-0.1in}    
    \caption{Generated examples of ablation study, with prompt "$\color{red}{S^*}$ is wearing superman costume".}
    \label{fig:ablation}
    \vspace{-0.2in}
\end{figure}

\paragraph{Fusion Sampling} First of all, we apply the proposed Fusion Sampling with both pre-trained and fine-tuned PromptNet. As shown in Figure \ref{fig:before_after_finetuning}, Fusion Sampling obtains better results on both pre-trained and fine-tuned models compared to baseline classifier-free sampling. We then investigate the effects of removing fusion stage or refinement stage in the proposed Fusion Sampling. As we can see from Figure \ref{fig:ablation}, removing refinement stage leads to the loss in detailed information, while removing fusion stage leads to a generated image with disorganized structure. Intuitively, $S^*$, which is the output of PromptNet, tries to generate a human face image following the structural information from the original image, while the text "is wearing superman costume" aims to generate a half-length photo. The conflicting nature of these two conditions results in an undesirable generation with a disorganized structure after we remove the fusion stage.

\begin{figure}[t!]
    \centering
    \includegraphics[width=0.73\linewidth]{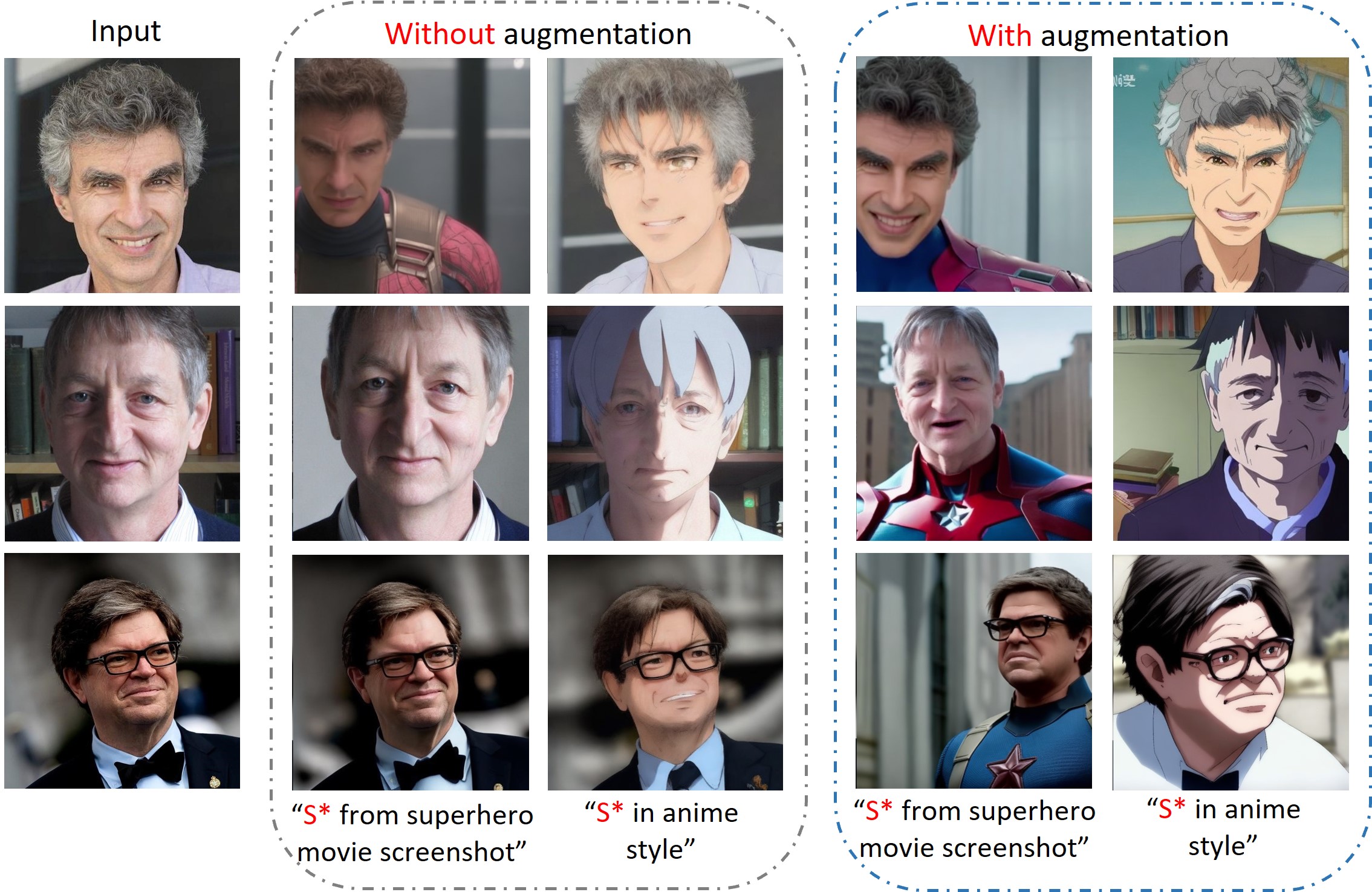}
    \vspace{-0.1in}
    \caption{Data augmentation in fine-tuning stage leads to performance improvement.}
    \label{fig:with_without_augmentation}
    \vspace{-0.2in}    
\end{figure}

\paragraph{Data Augmentation} We then analyze the effects of data augmentation. In particular, we conduct separate fine-tuning experiments: one with data augmentation and one without, both models are tested with Fusion Sampling after fine-tuning. The results are shown in Figure \ref{fig:with_without_augmentation}, we observe an improvement in performance as a result of employing data augmentation. Our data augmentation strategy is presented in the Appendix.

\section{Discussion}
Although the proposed framework has demonstrated remarkable capability in achieving high-quality customized generation, there are areas that can be improved. For instance, although ProFusion can reduce the
training time by only requiring a single training without the need of tuning regularization hyperparameters, the proposed Fusion Sampling actually results in an increased inference time. This is due to the division of each sampling step into two stages. In the future, we would like to explore ways to improve the efficiency of Fusion Sampling.

Similar to other related works, our framework utilizing large-scale text-to-image generation models can raise ethical implications, both positive and negative. On the one hand, customized generation can create images with sensitive information and spread misinformation; On the other hand, it also holds the potential to minimize model biases  as discussed in ~\cite{gal2022textualinversion,gal2023designing}. Thus it is crucial to exercise proper supervision when implementing these methods in real-world applications.

\section{Conclusion}
In this paper, we present ProFusion, a novel framework for customized generation. Different from related methods which employs regularization, ProFusion successfully performs customized generation without any regularization, thus exhibits superior capability for preserving fine-grained details with less training time.
Extensive experiments have demonstrated the effectiveness of the proposed ProFusion.

\bibliography{Reference.bib}
\bibliographystyle{plain}

\clearpage

\appendix

\section{More Generated examples}
\begin{figure}[h!]
    \centering
    \includegraphics[width=0.9\linewidth]{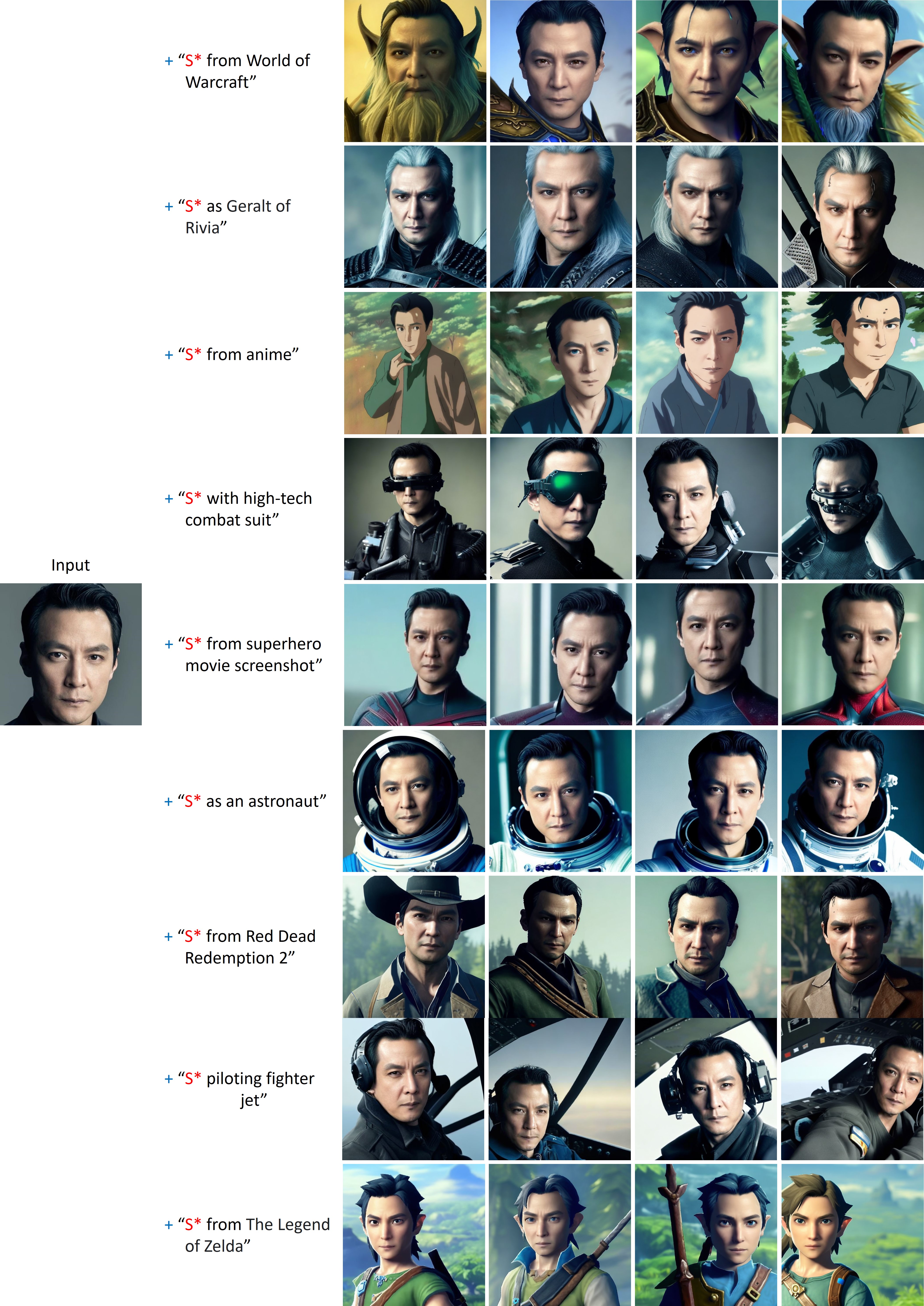}
    \caption{More results of customized generation with the proposed framework.}
    \label{fig:more_results_daniel}
\end{figure}

\begin{figure}[h!]
    \centering
    \includegraphics[width=0.75\linewidth]{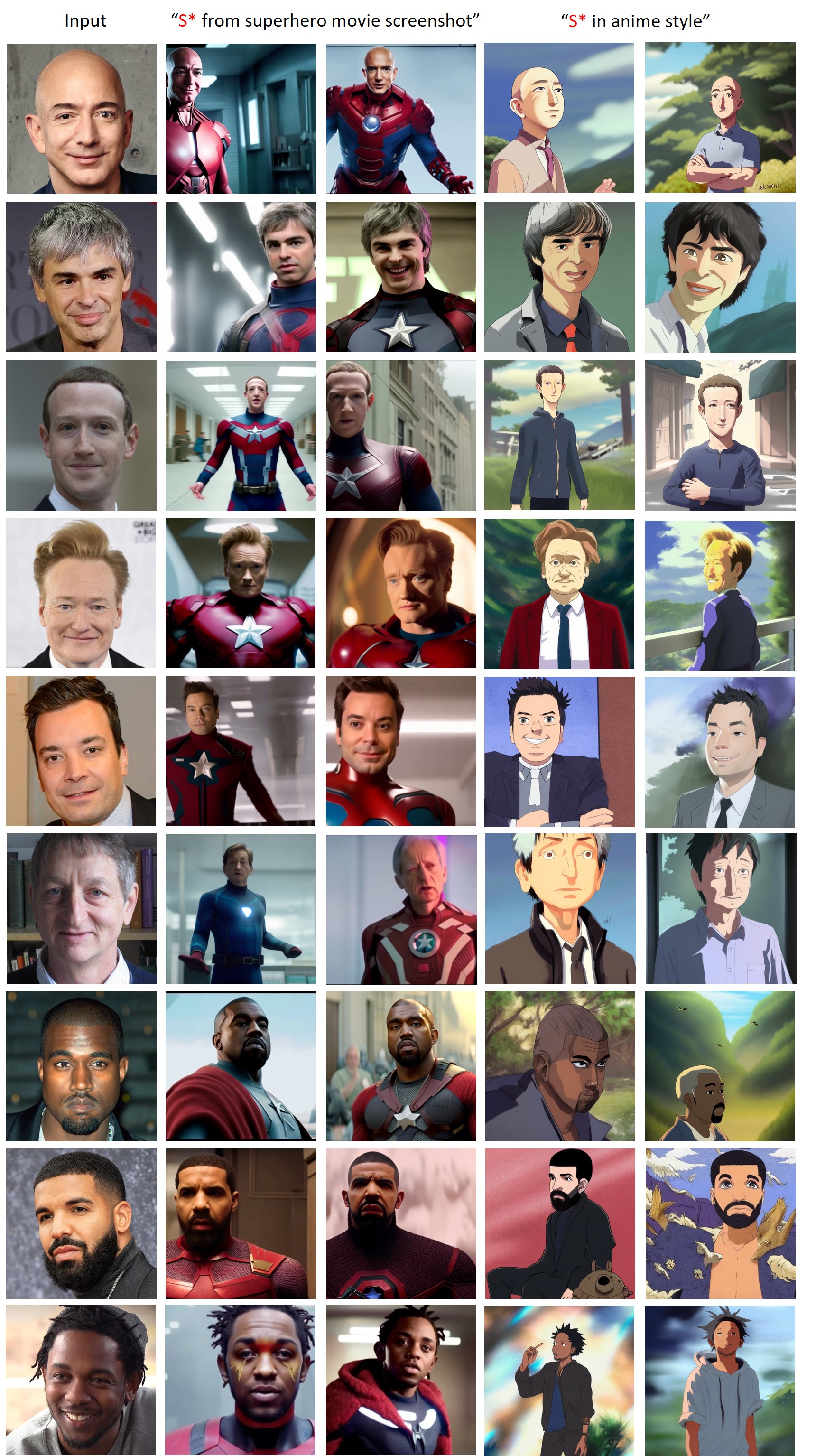}
    \caption{More results of  customized generation with the proposed framework.}
    \label{fig:more_results_1}
\end{figure}

\begin{figure}[h!]
    \centering
    \includegraphics[width=0.75\linewidth]{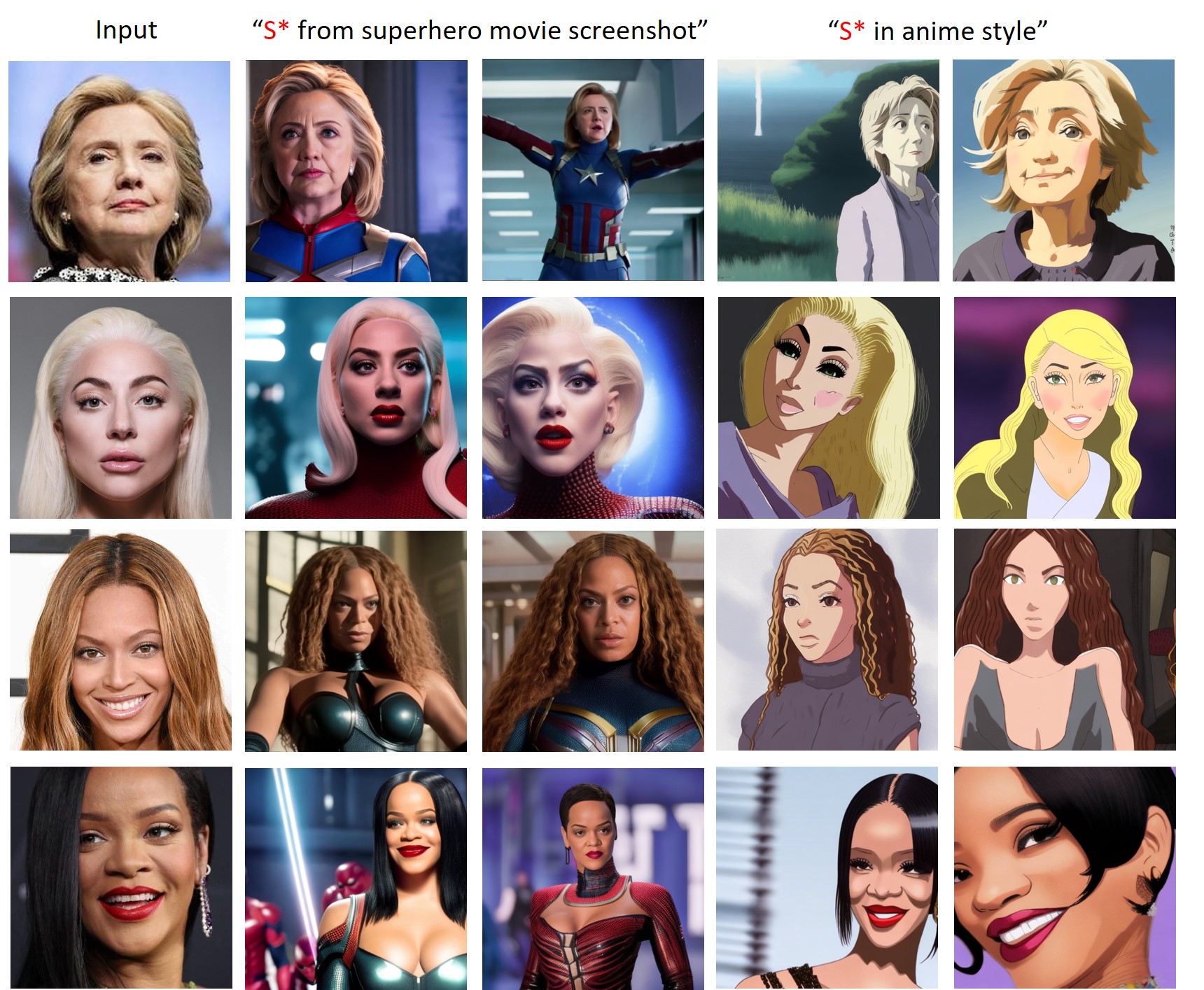}
    \caption{More results of  customized generation with the proposed framework.}
    \label{fig:more_results_2}
\end{figure}

Some more generated examples are provided in Figure \ref{fig:more_results_daniel}, Figure \ref{fig:more_results_1} and Figure \ref{fig:more_results_2}.
\clearpage

\clearpage

\section{Experiment Details}
\begin{figure}[t!]
    \centering
    \includegraphics[width=0.7\linewidth]{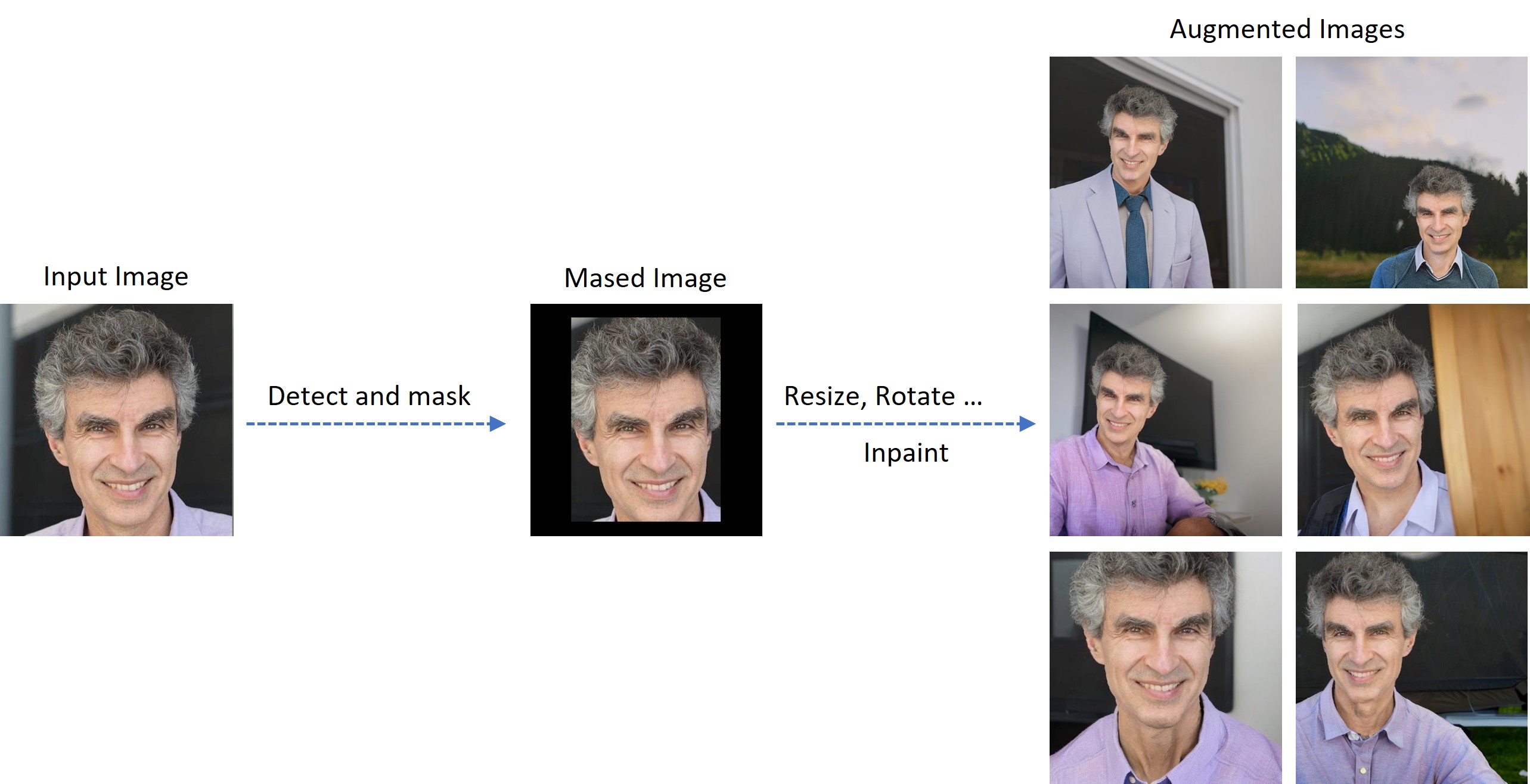}
    \caption{Illustration of data augmentation in our fine-tuning stage.}
    \label{fig:augmentation}
\end{figure}

We provide some experiment details in this section.
\paragraph{Data augmentation} We implement data augmentation at the fine-tuning stage, which is illustrated in Figure \ref{fig:augmentation}. Given a testing image, we first create a masked image where only the target face/object is unmasked. The masked image will be fed into a pre-trained Stable Diffusion inpainting model after random resize and rotation. The inpainting model will generate multiple augmented images, with different background. We use a postive prompt "a photo of a man/woman, highly detailed, soft natural lighting, photo realism, professional portrait, ultra-detailed, 4k resolution, wallpaper, hd wallpaper." and a negative prompt "magazine, frame, tiled, repeated, multiple people, multiple faces, group of people, split frame, multiple panel, split image, watermark, boarder, diptych, triptych" with a classifier-free guidance of 7.5 during inpainting. 

\paragraph{PromptNet}
Our PromptNet is an encoder contains 5 encoder blocks, which are similar to the downsize and middle blocks in Stable Diffusion 2. The parameters are intialized with value from pre-trained Stable Diffusion 2 when applicable. Different from the blocks in Stable Diffusion 2, we use image embeddings  from pre-trained CLIP ViT-H/14 instead of text embeddings as the contents for cross attention layers. The inputs $\bar{\xb}_0$ and $\xb_t$ are first processed by different convolution layers, whose outputs are summed to serve as the input for the following blocks.

\paragraph{Human Evaluation}
Due to the fact that we do not have official implementation and pre-trained models of E4T~\cite{gal2023designing}, we directly take some generated examples from their paper for fair comparison. Then we use corresponding prompts in our framework to generate images to be compared. 
Specifically, there are 39 source image and prompt pairs for five different methods and each generated image is evaluated by five different workers with expertise. These workers are all from the US and required to have performed at least 10,000 approved assignments with an approval rate $\geq$ 98\%. The human evaluation user interface is shown in Figure \ref{fig:human_evaluation}.

\begin{figure}[t!]
    \centering
    \includegraphics[width=\linewidth]{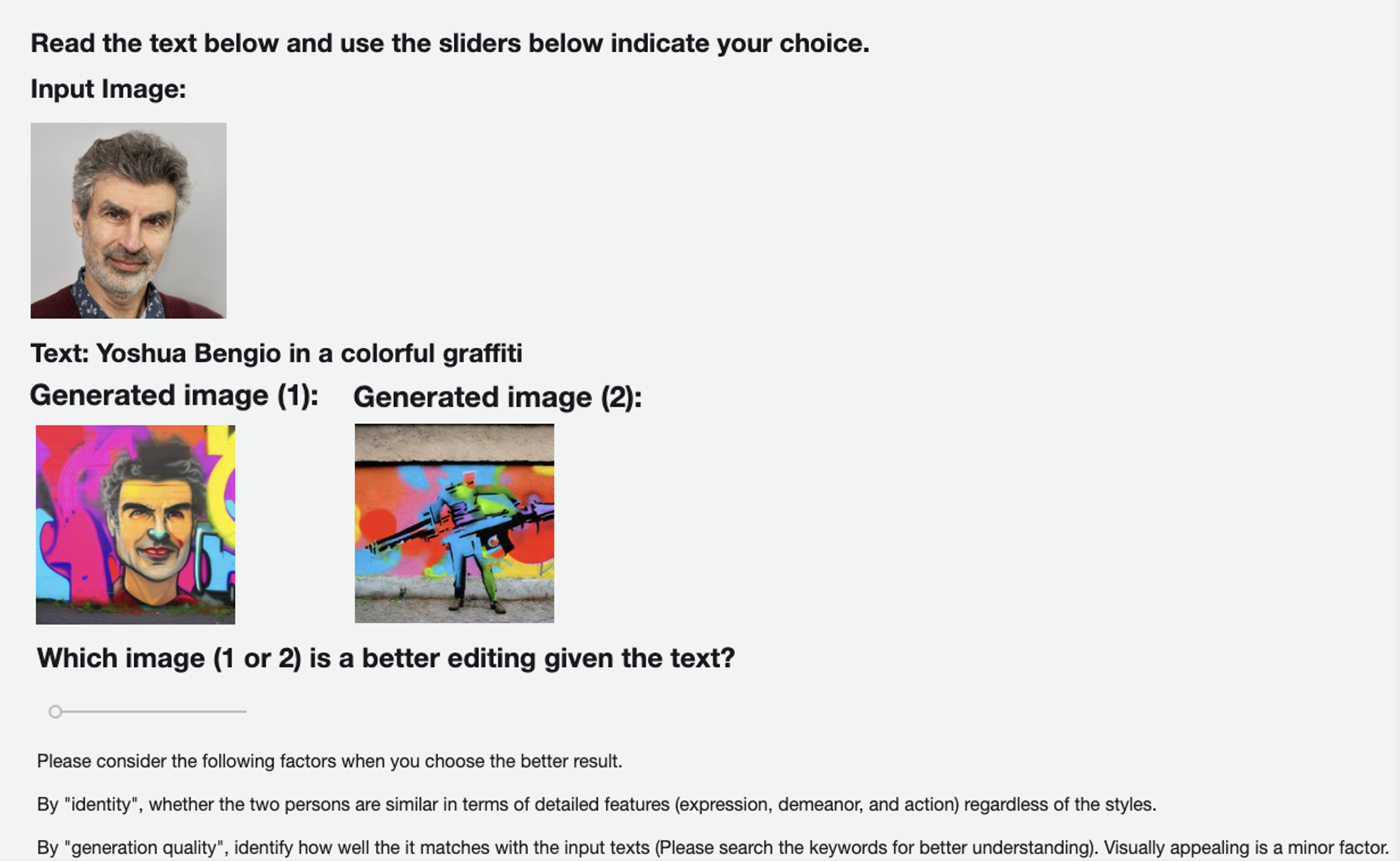}
    \caption{Human Evaluation User Interface.}
    \label{fig:humaneval}
\end{figure}


\end{document}